# Polish to English Statistical Machine Translation

Krzysztof Wołk, *Polish Japanese Institute of Technology*

## Abstract

This research explores the effects of various training settings on a Polish to English Statistical Machine Translation system for spoken language. Various elements of the TED, Europarl, and OPUS parallel text corpora were used as the basis for training of language models, for development, tuning and testing of the translation system. The BLEU, NIST, METEOR and TER metrics were used to evaluate the effects of the data preparations on the translation results.

## 1. Introduction

Statistical Machine Translation (SMT) systems use large, parallel, bilingual corpora of texts to build a model of both source and target natural languages. The model is developed based on statistics derived from the corpus. Automated machine translation is a considerable challenge, requiring a very deep understanding of the text or sufficient data for statistical modeling [1].

Polish is one of the most complex West-Slavic languages, which represents a serious challenge to any SMT system. The grammar of the Polish language, with its complicated rules and elements, together with a big vocabulary (due to complex declension) are the main reasons for its complexity. This greatly affects the data and data structure required for statistical models of translation. The lack of available and appropriate resources required for data input to SMT systems presents another problem. SMT systems should work best in specified, not too wide text domains and will not perform well for a general use. Good quality parallel data especially in a required domain has low availability. Polish and English differ also in syntax. English is a positional language, which means that the syntactic order (the order of words in a sentence) plays a very important role, particularly due to very limited inflection of words (e.g. lack of declension endings). Sometimes, the position of a word in a sentence is the only indicator of the sentence meaning. In the English sentence, the subject group comes before the predicate, so the sentence is ordered according to the Subject-Verb-Object (SVO) schema. In Polish, however, there is no specific word order imposed and the word order has no decisive influence on the understanding of the sentence. One can express the same thought in several ways, which is not possible in English. For example, the sentence „I bought myself a new car." can be written in Polish as „Kupiłem sobie nowy samochód", or "Nowy samochód sobie kupiłem.", or "Sobie kupiłem nowy samochód.", or „Samochód nowy sobie kupiłem." Differences in potential sentence orders make the translation process more complex, especially when working on a phrase-model with no additional lexical information.

As a result, the progress of SMT systems for Polish is substantially slower as compared to other languages. The aim of this work is to prepare an SMT system for translation from Polish to English to address the TED [2] task requirements. Another objective was to create SMT systems for the Europarliament speeches and for the Open Parallel Corpus (OPUS) movie subtitle data set. Those data sets are representative samples of the spoken language, which differs from the written one. Thus, the aim is also to test if SMT methods are suitable also for this kind of language.

This paper is structured as follows: Section 2 explains the Polish data preparation. Section 3 presents the English language issues. Section 4 describes the translation evaluation methods. Section 5 discusses the results. Sections 6 and 7 discuss potential implications and future work.

## 2. Preparation of the Polish data

The Polish data in the TED talks (15 MB) include almost 2 million words that are not tokenized. The transcripts themselves are provided as pure text encoded with UTF-8 and the transcripts are prepared by the FBK team [3]. In addition, they are separated into sentences (one per line) and aligned in language pairs. A substantial amount (90 MB) of the English data includes the PL-EN Europarl v7 corpus prepared in Euromatrix project [4]. We also used the

500 MB OPUS corpus created from a collection of documents obtained from movie subtitles. [5]

It should be emphasized that both automatic and manual preprocessing of this training information was required. The extraction of the transcription data from the provided XML files ensured an equal number of lines for English and Polish. However, some of the discrepancies in the text parallelism could not be avoided. These discrepancies are mainly repetitions of the Polish text not included in the English text.

A number of English names are present in the Polish text. There are also some words from other languages (e.g., German and French). Some translations are incorrect.

The size of the vocabulary is 88,158 Polish unique words and 41,684 English unique words. The disproportionate vocabulary sizes are also a challenge.

Before the use of a training translation model, preprocessing that included removal of long sentences (set at 80 characters) had to be performed. The Moses toolkit scripts[6] were used for this purpose. Moses is an open-source toolkit for statistical machine translation which supports linguistically motivated factors, confusion network decoding, and efficient data formats for translation models and language models. In addition to the SMT decoder, the toolkit also includes a wide variety of tools for training, tuning and applying the system to many translation tasks. In addition, the text in the TED data set had to be repaired in a number of ways to correct: spelling errors, grammar errors, ensuring only one sentence on each line, removal of language translations not of interest, removal of HTML and XML tags within text, removal of strange symbols not existing in a specific language and repetitions of words and sentences.

The final processing covers 134,678 lines from the Polish to English corpus. However, the disproportionate vocabulary sizes remain, with 41,163 English words and 85,775 Polish words. One of the solutions to this problem (according to work of Bojar [7]) was to use stems instead of surface forms that reduced the Polish vocabulary size to 40,346. Such a solution also requires a creation of an SMT system from Polish stems to plain Polish. Subsequently, morphosyntactic tagging, using the Wroclaw Natural Language Processing (NLP) tools, was included as an additional information source for the SMT system preparation. It can be also used as a first step for implementing a factored SMT system that, unlike a phrase-based system, includes morphological analysis, translation of lemmas and features as well as generation of surface forms. Incorporating additional linguistic information should effectively improve translation performance. [8]

## 2.1.   Polish stem extraction

As previously mentioned, stems extracted from Polish words are used instead of surface forms to overcome the problem of the huge difference in vocabulary sizes. Keeping in mind that the target language was English in the form of normal sentences, it was not necessary to introduce models for converting the stems to the appropriate grammatical forms. For Polish stem extraction, a set of natural language processing tools available at http://nlp.pwr.wroc.pl was used [9]. These tools can be used for:

a)      Tokenization
b)      Morphosyntactic analysis
c)      Shallow parsing as chunking
d)      Text transformation into the featured vectors

The following two components are also included:

1)      MACA –a universal framework used to connect the different morphological data
2)      WCRFT – this framework combines conditional random fields and tiered tagging

These tools used in sequence provide an XML output. It includes the surface form of the tokens, stems and morphosyntatic tags. An example of such data is showed in section 2.2.

## 2.2.   Morphosyntactic element tagging with standard tools

Wroclaw's tools were used to tag morphosyntactic elements. More precise tagging can be achieved with these settings. In addition, every tag in this tagset consists of specific grammatical classes with specific values for particular attributes. Furthermore, these grammatical classes include attributes with values that require additional specification. For example, nouns require numbers while adverbs require an appropriate degree of an attribute. This causes segmentation of the input data, including tokenization of the words in a different way as compared to the Moses tools. On the other hand, this causes problems with building parallel corpora. This can be solved by placing markers at the end of input lines.

In the following example, where pl.gen. "men" is derived from sin.nom."człowiek" (*man*) or pl.nom. "ludzie" (*people*), it can be demonstrated how one tag is used where, in the most difficult cases, more possible tags are provided.

```
<tok>
<orth>ludzi</orth>
<lex disamb="1"> <base>człowiek</base>
<ctag>subst:pl:gen:m1</ctag></lex>
<lex disamb="1"> <base>ludzie</base>
<ctag>subst:pl:gen:m1</ctag></lex>
</tok>
```

In this example, only one form (the first stem) is used for further processing.

I developed an XML extractor tool to generate three different corpora for the Polish language data: (a) words in the infinitive form, (b) a Subject-Verb-Object (SVO) word order, and (c) both the infinitive form and the SVO word order. This allows experiments with those preprocessing techniques.

Moreover, some of the NLP tools use the Windows-1250 Eastern Europe Character Encoding, which requires a conversion of information to and from the UTF-8 encoding
that is commonly used in other, standard tools.

## 3. English Data Preparation

The preparation of the English data was definitively less complicated than for Polish. I developed a tool to clean the English data by removing foreign words, strange symbols, etc.

## 4. Evaluation Methods

Metrics are necessary to measure the quality of translations produced by the SMT systems. For this, various automated metrics are available to compare SMT translations to high quality human translations. Since each human translator produces a translation with different word choices and orders, the best metrics measure SMT output against multiple reference human translations. Among the commonly used SMT metrics are: Bilingual Evaluation Understudy (BLEU), the U.S. National Institute of Standards & Technology (NIST) metric, the Metric for Evaluation of Translation with Explicit Ordering (METEOR), and Translation Error Rate (TER). These metrics will now be briefly discussed. [10]

BLEU was one of the first metrics to demonstrate high correlation with reference human translations. The general approach for BLEU, as described in [9], is to attempt to match variable length phrases to reference translations. Weighted averages of the matches are then used to calculate the metric. The use of different weighting schemes leads to a family of BLEU metrics, such as the standard BLEU, Multi-BLEU, and BLEU-C. [11]

As discussed in [11], the basic BLEU metric is:

$$BLEU = P_B \exp\left(\sum_{n=0}^{N} w_n \log p_n\right)$$

where $p_n$ is an $n$-gram precision using $n$-grams up to length N and positive weights $w_n$ that sum to one. The brevity penalty $P_B$ is calculated as:

$$P_B = \begin{cases} 1, & c > r \\ e^{(1-r/c)}, & c \leq r \end{cases}$$

where $c$ is the length of a candidate translation, and $r$ is the effective reference corpus length. [9]

The standard BLEU metric calculates the matches between $n$-grams of the SMT and human translations, without considering position of the words or phrases within the texts. In addition, the total count of each candidate SMT word is limited by the corresponding word count in each human reference translation. This avoids bias that would enable SMT systems to overuse high confidence words in order to boost their score. BLEU applies this approach to texts sentence by sentence, and then computes a score for the overall SMT output text. In doing this, the geometric mean of the individual scores is used, along with a penalty for excessive brevity in translation. [9]

The NIST metric seeks to improve the BLEU metric by valuing information content in several ways. It takes the arithmetic versus geometric mean of the $n$-gram matches to reward good translation of rare words. The NIST metric also gives heavier weights to rarer words. Lastly, it reduces the brevity penalty when there is a smaller variation in translation length. This metric has demonstrated that these changes improve the baseline BLEU metric. [12]

The METEOR metric, developed by the Language Technologies Institute of Carnegie Mellon University, is also intended to improve the BLEU metric. METEOR rewards recall by modifying the BLEU brevity penalty, takes into account higher order $n$-grams to reward matches in word order, and uses arithmetic vice geometric averaging. For multiple reference translations, METEOR reports the best score for word-to-word matches. Banerjee and Lavie [13] describe this metric in detail.

As found in [13], this metric is calculated as follows:

$$METEOR = \left(\frac{10\,P\,R}{R + 9\,P}\right)(1 - P_M)$$

where $P$ is the unigram precision and $R$ is the unigram recall. The METEOR brevity penalty $P_M$ is:

$$P_M = 0.5 \left(\frac{C}{M_U}\right)$$

where $C$ is the minimum number of chunks such that all unigrams in the machine translation are mapped to unigrams in the reference translation. $M_U$ is the number of unigrams that matched.

The METEOR metric incorporates a sophisticated word alignment technique that works incrementally. Each alignment stage attempts to map previously unmapped words in the SMT and reference translations. In the first phase of each stage, METEOR attempts three different types of word-to-word mappings, in the following order: exact matches, matches using stemming, and matches of synonyms. The second stage uses the resulting word mappings to evaluate word order similarity. [13]

Once a final alignment of the texts is achieved, METEOR calculates precision similar to the way the NIST metric calculates it. METEOR also calculates word-level recall between the SMT translation and the references, and combines this with precision by computing a harmonic mean that weights recall higher than precision. Lastly, METEOR penalizes shorter $n$-gram matches and rewards longer matches. [13]

TER is one of the most recent and intuitive SMT metrics developed. This metric determines the minimum number of human edits required for an SMT translation to match a reference translation in meaning and fluency. Required human edits might include inserting words, deleting words, substituting words, and changing the order or words or phrases. [14]

As given in [14], this metric is given by:

$$TER = \frac{E}{w_R}$$

where $E$ is the number of human translator edits of the machine translation required such that it matches the closest reference translation. $w_R$ is the average length of the references.

## 5. Experimental Results

A number of experiments has been performed to evaluate various versions for our SMT system, translating three sets of texts between English and Polish: TED, Europarl, and the Open Parallel Corpus (OPUS). In general, such experiments require parallel corpora with four types of files: a large parallel file to train the SMT system, a large language model of the target language (containing statistically assigned probabilities to sequences of words by means of probability distribution), small files used to tune the SMT system during development (DEV), and small files used for testing (TEST).

The experiments involved a number of steps. Processing of the corpora was accomplished, including tokenization, cleaning, factorization, conversion to lower case, splitting, and a final cleaning after splitting. Training data was processed, and the language model was developed. Tuning was performed for each experiment. Lastly, the experiments were conducted.

The testing was done using the Moses open source SMT toolkit with its Experiment Management System (EMS) [15]. The SRI Language Modeling Toolkit (SRILM) [16] with an interpolated version of the Kneser-Key discounting (interpolate –unk –kndiscount) was used for 5-gram language model training. We used the MGIZA++ tool for word and phrase alignment. MGIZA++ is a multi-threaded version ofthe well-known GIZA++ tool [17]. The symmetrization method was set to grow-diag-final-and for word alignment processing. Two-way direction alignments obtained from GIZA++ are firstly intersected, so only the alignment points that occur in both alignments remained. In second phase additional alignment points existing in their union are added. The growing step adds potential alignment points of unaligned words and neighbors. Neighborhood can be set directly to left, right, top or bottom, as well as to diagonal (grow-diag). In the final step, alignment points between words from which at least one is unaligned are added (grow-diag-final). If the grow-diag-final-and method is used, an alignment point between two unaligned words appears[18]. KenLM [19] was used to binarize the language model, with a lexical reordering set to use the msd-bidirectional-fe model. Reordering probabilities of phrases are conditioned on lexical values of a phrase. It considers three different orientation types on source ant target phrases like monotone(M), swap(S) and discontinuous(D). The bidirectional reordering model adds probabilities of possible mutual positions of source counterparts to current and following phrases. Probability distribution to a foreign phrase is determined by "f" and to the English phrase by "e".[20,21]

First, Polish-to-English translation experiments were conducted using the TED corpus with variations in the data used for training language model, tuning during development, and finally testing. The following training preparations were varied in the experiments: original TED data with no preparation, purely automated cleaning (A) of the data set, a combination of automated and then manual cleaning (A/M), cleaning with punctuation (P) removed, cleaning with verbs transformed to their infinitive (INF) form, cleaning and conversion to SVO word order, training without use of selected test data, and combinations of these preparations.

The Language Models (for English) used in the experiments were either the original TED model, or one that had been automatically cleaned. The experiments used a variety of the TST2010, TST2011, and DEV2010 data sets, in their original and modified forms, to tune the translation output during development. The modified forms used include: A/M cleaning, verbs changed to their INF form, SVO order normalization, and combinations of these adjustments. Lastly, the experiments utilized the TST2010 and TST2011 files with similar differences in preparation as for the development/tuning data.

The TED experiments are defined in Table 1, measured by the BLEU, NIST, METEOR (MET), and TER metrics. Note that a lower value of the TER metric is better, while the other metrics are better when their values are higher. Experiment 0 stands for baseline system with no improvements. The highest quality gain was obtained by cleaning and repairing mistakes in train, development and test data. Perfection in data quality proved to be key to good translation, however it must be noted that originally test data sets were also present in training data. Another big improvement is related to test number 14 where infinitive forms were used. Test no. 16 shows that converting word order to SVO did not make a positive impact.

Additional experiments were conducted with the EUROPARL parallel corpus. Six experiments examined the effect of variations on Polish-to-English (PL-EN) SMT. The training data sets used the original data without modification (Experiment 0), automatically cleaned data, or cleaned data with foreign words (F) removed. The language model variations used in these experiments included: unaltered language model data, automatically cleaned data (Ex. 2), foreign words removed(Ex. 3), cleaned data with unknown words (UW) removed, extraction of English from the French-English (FR-EN) data, and combinations of these modifications. Cleaning slightly improved quality, however more improvement was obtained by extending the language model in Experiment 5. One experiment involving English-to-Polish (Experiment 6) translation (using automatically cleaned data with foreign words removed, and the original EUROPARL language model) was performed for comparison purposes. In all these experiments, the development/tuning and test data were randomly generated. The EUROPARL experiments are defined in Table 2.

Two experiments were performed using the OPUS data set to compare the translation of Polish to English with that of English to Polish. These experiments used the original OPUS training and language model data sets, and randomly generated development/tuning and testing data sets. Table 3 defines the experiments. Both resulted in a similar score, most likely because these data sets consist of short and easy phrases.

# 6. Discussion

Several conclusions can be drawn from the experimental results presented here. Automatic and manual cleaning of the training files had the most impact, among the variations examined, on improving translation performance. In particular, automatic cleaning and conversion of verbs to their infinitive forms improved translation performance the most. This is likely due to reduction of the Polish vocabulary size.

Automatic cleaning of data used for the Language Model improved translation performance in the experiments with the EUROPARL data set. Use of a larger corpus for language model training improves the translation scores. In particular, Experiment 6th in the EUROPARL experiments, where two language models were combined into one, resulted in high scores.

A limited number of experiments with the OPUS data set indicate similar performance when translating from English to Polish and from Polish to English. However, it should be noted that OPUS has very large amounts of data and is composed of easy, short movie subtitles.

**TED Polish-to-English Experiments with Results**

| # | Training | Lang. Model | Tuning | Test | BLEU | NIST | MET | TER |
|---|----------|-------------|--------|------|------|------|-----|-----|
| 0 | Original | Original | **DEV2010** | **TST2010** | **15.83** | **5.23** | **49.11** | **67.15** |
| 1 | **A/M** | **Original** | **DEV2010** | **TST2010** | **67.19** | **10.93** | **82.39** | **27.03** |
| 2 | A/M No P | Original | DEV2010 | TST2010 | 57.28 | 9.89 | 76.44 | 34.71 |
| 3 | A | Original | DEV2010 | TST2010 | 53.36 | 9.74 | 75.47 | 35.62 |
| 4 | A | A | DEV2010 | TST2010 | 52.88 | 9.73 | 75.29 | 35.61 |
| 5 | A/M No P | A | DEV2010 | TST2010 | 56.59 | 9.85 | 76.58 | 34.91 |
| 6 | A/M | A | DEV2010 | TST2010 | 66.46 | 10.83 | 81.85 | 27.67 |
| 7 | A/M | A | DEV2010 | TST2010 A/M | 66.72 | 10.87 | 81.85 | 27.54 |
| 8 | A/M | A | DEV2010 A/M | TST2010 A/M | 67.32 | 10.93 | 82.46 | 26.90 |
| 9 | A/M No TST2010 | Original | DEV2010 A/M | TST2010 A/M | 61.22 | 10.55 | 79.44 | 31.30 |
| 10 | Original | Original | DEV2010 | TST2010 | 19.29 | 5.69 | 52.58 | 63.41 |
| 11 | A/M | Original | TST2011 A | TST2011 A/M | 17.37 | 5.72 | 53.23 | 61.92 |
| 12 | Original | Original | TST2011 | TST2011 | 19.58 | 5.76 | 52.92 | 62.47 |
| 13 | A/M INF | Original | DEV2010 | TST2010 INF | 67.99 | 11.64 | 84.67 | 23.74 |
| 14 | **A/M INF** | **Original** | **DEV2010 INF** | **TST2010 INF** | **75.78** | **11.94** | **85.82** | **21.89** |
| 15 | A/M SVO | Original | DEV2010 | TST2010 SVO | 59.95 | 10.99 | 80.55 | 30.36 |
| 16 | **A/M SVO** | **Original** | **DEV2010 SVO** | **TST2010 SVO** | **62.10** | **10.46** | **78.34** | **34.68** |
| 17 | A/M SVO + INF | Original | DEV2010 | TST2010 SVO + INF | 52.23 | 9.31 | 72.09 | 43.30 |
| 18 | A/M SVO + INF | Original | DEV2010 SVO + INF | TST2010 SVO + INF | 59.74 | 10.21 | 76.44 | 36.86 |

## 7. Future Work

Several potential avenues for future work are of interest. Additional experiments using extended language models are warranted to determine if this improves SMT scores. We are also interested in developing a web crawler to obtain additional movie subtitles for the OPUS data set. In general, obtaining more Polish language data would likely prove useful.

The majority of the experiments conducted involved Polish-to-English translation. It would be of interest to conduct additional English-to-Polish experiments, especially using the TED data set, for comparison. Increased cleaning automation is desired. Further investigation of infinitive preparation effects would also be valuable. Lastly, cleaning of the OPUS and EUROPARL data would be very beneficial. However, this would be a daunting task, since each language has approximately 500MB size text files.

Converting Polish verbs to infinitives reduces the Polish vocabulary, which should improve English to Polish translation performance. Polish to English translation typically outscores English to Polish translation, even on the same data. This requires further evaluation.

An ideal SMT system should be fully automatic. To zuse infinitives, we will have to make this conversion automatic with usage of Wroclaw NLP tools. Lastly, it is our objective to create two SMT systems, one converting Polish verbs to Polish infinitives, and another converting Polish infinitives to English in order to make translations fully automatic.

## 8. Conclusions

The analysis of our experiments led us to conclude that the results of the translations, in which the BLEU measure is greater than 70, can be considered satisfactory within the text domain. This high level of evaluation score should be understandable without problems for a human and good enough to help him in his work. Such systems can be used in practical applications, also when it comes to Polish translations. It may be particularly helpful with the Polish language, which is one of the most complex in terms of its structure, grammar and spelling.

## 9. Acknowledgements

This work is supported by the European Community from the European Social Fund within the Interkadra project UDA-POKL-04.01.01-00-014/10-00 and Eu-Bridge 7th FR EU project (grant agreement n°287658).

Table 2.

**EUROPARL Translation Experiments with Results**

| # | Trans. | Training | Lang. Model | Tuning | Test | BLEU | NIST | MET | TER |
|---|--------|----------|-------------|--------|------|------|------|-----|-----|
| **0** | **PL-EN** | **Original** | **Original** | **Random** | **Random** | **81.65** | **12.62** | **92.04** | **15.09** |
| 1 | PL-EN | A | A | Random | Random | 83.61 | 12.81 | 92.51 | 13.57 |
| 2 | PL-EN | A No F | A No F | Random | Random | 82.35 | 12.69 | 92.00 | 14.39 |
| 3 | PL-EN | A No F | FR-EN Train. Data | Random | Random | 82.69 | 12.77 | 92.27 | 13.91 |
| 4 | PL-EN | A No F | FR-EN Train. Data Norm. | Random | Random | 81.78 | 12.67 | 91.86 | 14.52 |
| 5 | **PL-EN** | **A No F** | **Original + FR-EN Train. Data** | **Random** | **Random** | **86.16** | **13.02** | **93.13** | **12.20** |
| 6 | **EN-PL** | **A No F** | **Original** | **Random** | **Random** | **80.64** | **12.22** | **88.24** | **16.73** |

Table 3.

**OPUS Translation Experiments with Results**

| # | Trans. | Training | Lang. Model | Tuning | Test | BLEU | NIST | MET | TER |
|---|--------|----------|-------------|--------|------|------|------|-----|-----|
| 0 | PL-EN | Original | Original | Random | Random | 64.56 | 8.29 | 72.76 | 35.73 |
| 1 | EN-PL | Original | Original | Random | Random | 65.97 | 8.72 | 74.77 | 36.28 |

**Work Address**

Mgr inż. Krzysztof Wołk
kwolk@pjwstk.edu.pl
Multimedia Department
Polish Japanese Institute of Technology
ul. Koszykowa 86
02-008 Warszawa
tel. (22) 58 44 500
Fax. (22) 58 44 501